\pdfoutput=1
\documentclass[runningheads]{llncs}

\usepackage[T1]{fontenc}
\usepackage{amsmath}
\usepackage{graphicx}
\usepackage{multirow}
\usepackage{svg}
\usepackage{makecell}
\usepackage{amssymb}

\DeclareMathOperator{\Diag}{Diag}
\DeclareMathOperator{\ZeroDiag}{ZeroDiag}
\DeclareMathOperator{\Heaviside}{\Theta}
\DeclareMathOperator{\Conv}{Conv}
\DeclareMathOperator{\Sigmoid}{\sigma}

\begin{document}
\title{A Cross-Lingual Meta-Learning Method Based on Domain Adaptation for Speech Emotion Recognition}
\titlerunning{A Cross-Lingual Meta-Learning Method Based on DANN for SER}

\author{David-Gabriel Ion\textsuperscript{1} \and
Răzvan-Alexandru Smădu\textsuperscript{1} \and
Dumitru-Clementin Cercel\textsuperscript{1,\rm *}  \and
Florin Pop\textsuperscript{1,2,3} \and
Mihaela-Claudia	Cercel\textsuperscript{4,5}
}

\institute{Faculty of Automatic Control and Computers, National University of Science and Technology POLITEHNICA Bucharest, Bucharest Romania \and
 National Institute for Research and Development in Informatics - ICI Bucharest, Bucharest, Romania \and
 Academy of Romanian Scientists, Bucharest, Romania \and
 Paris 1 Panthéon-Sorbonne University, Paris, France \and
 University of Bucharest, Bucharest, Romania}

\authorrunning{D.-G. Ion et al.}

\def\thefootnote{*}\footnotetext{Corresponding author: dumitru.cercel@upb.ro.}
\def\thefootnote{\arabic{footnote}}
%
%
\maketitle              
\begin{abstract}
Best-performing speech models are trained on large amounts of data in the language they are meant to work for. However, most languages have sparse data, making training models challenging. This shortage of data is even more prevalent in speech emotion recognition. Our work explores the model's performance in limited data, specifically for speech emotion recognition. Meta-learning specializes in improving the few-shot learning. As a result, we employ meta-learning techniques on speech emotion recognition tasks, accent recognition, and person identification. To this end, we propose a series of improvements over the multistage meta-learning method. Unlike other works focusing on smaller models due to the high computational cost of meta-learning algorithms, we take a more practical approach. We incorporate a large pre-trained backbone and a prototypical network, making our methods more feasible and applicable. Our most notable contribution is an improved fine-tuning technique during meta-testing that significantly boosts the performance on out-of-distribution datasets. This result, together with incremental improvements from several other works, helped us achieve accuracy scores of 83.78\% and 56.30\% for Greek and Romanian speech emotion recognition datasets not included in the training or validation splits in the context of 4-way 5-shot learning.

\end{abstract}

\section{Introduction}


Speech emotion recognition presents significant challenges regarding available data, which is crucial in obtaining a high-performance model. Data availability is a general issue even for speech. Common Voice \cite{CommonVoice}, one of the largest publicly available multilingual datasets, only includes 43 languages with more than 100 hours of data, 11 of which have more than 1,000 hours.


Given this, we focus on the low-data scenario, where techniques such as meta-learning \cite{finn2017model} are suitable. Meta-learning \cite{hospedales2021meta} encompasses a set of algorithms and techniques that focus on learning how to learn, essentially training models that can generalize on unknown but related data distributions. It is particularly useful for one-shot and few-shot learning.

This paper focuses on the practical aspect, similar to Hu et al. \cite{hu2022pushing}; hence, our work starts from those findings. The authors defined three distinct steps to achieve a high-performance pipeline: pre-training, meta-learning, and fine-tuning, namely P\textbf{$>$}M\textbf{$>$}F. Pre-training usually refers to a self-supervised training step that yields a large pre-trained model. Meta-learning turns our pre-trained backbone into a meta-learner. Finally, in fine-tuning, we train our meta-learner for a few steps on the newly seen data to squeeze better results out of our model at the expense of computational costs.

Our contribution consists of improving each step of this multistage process in at least some regard. We employ a larger backbone structure for the pre-training stage featuring the Wav2Vec2 XLS-R 300M transformer model \cite{babu2021xlsr,baevski2020wav2vec,NIPS2017_3f5ee243} instead of other works \cite{cai2020meta,feng2021few} that employ smaller models.
The meta-learning stage is the focus point of our research. We employ a prototypical network \cite{snell2017prototypical} on top of our model, chosen for its simplicity, performance, and versatility \cite{hu2022pushing}. We use a feature extractor network as an intermediary between our transformer model and the prototypical network. This network aims to reduce the variable-length matrix outputted by the transformer into a fixed-length vector, which acts as the sample prototype in our meta-learning algorithm. 
Finally, in the fine-tuning stage, we test the P\textbf{$>$}M\textbf{$>$}F method \cite{hu2022pushing} and then introduce a novel approach that is twice as fast and obtains a higher accuracy in the context of few-shot learning with minimal data augmentation.

To summarize, the main contributions of this work are as follows:
\begin{itemize}
    \item Introduction of a novel meta-testing fine-tuning technique that is both faster and yields better performance;
    \item Analysis of various feature extractors used to produce embeddings employed in the prototypical network;
    \item A study into the performance achieved in speech emotion recognition while training on person identification and accent recognition.
\end{itemize}

\section{Related Work}

Feng et al. \cite{feng2021few} introduced a setting employing a siamese network \cite{koch2015siamese} as their meta-learning algorithm. Their method used a more traditional approach, with extensive data preprocessing and a smaller model. In contrast, our method leverages the power of a large pre-trained model, namely Wav2Vec2 XLS-R 300M.

Chopra et al. \cite{chopra2021meta} made a significant stride by employing an optimization-based model-agnostic meta-learning algorithm, MAML \cite{finn2017model}. Their work involved training on the entire test dataset instead of only 4 to 20 samples, which substantially improved per-sample efficiency. However, our prototypical network would not be suitable for their use case.

Cai et al. \cite{cai2020meta} explored a similar approach by adding a two-stage training process. In the first stage, a model was trained on the valence-activation-dominance multi-label classification problem. During the next stage, another model was initialized with the weights resulting from the previous step before starting the training process.

Liu et al. \cite{liu2023speech} employed domain adaptation \cite{ganin2015unsupervised} to solve the speech emotion recognition problem. It featured a two-step training process, similar to \cite{cai2020meta}, with the difference that the model was initially trained on the source domain instead of a different multi-label classification task on the target domain. With this, the second stage, the actual meta-learning, can begin with part of the model having the weights already partially trained.

\section{Methodology}

\subsection{Architecture Overview}

\begin{figure}[t]
\centering
\includegraphics[width=\textwidth]{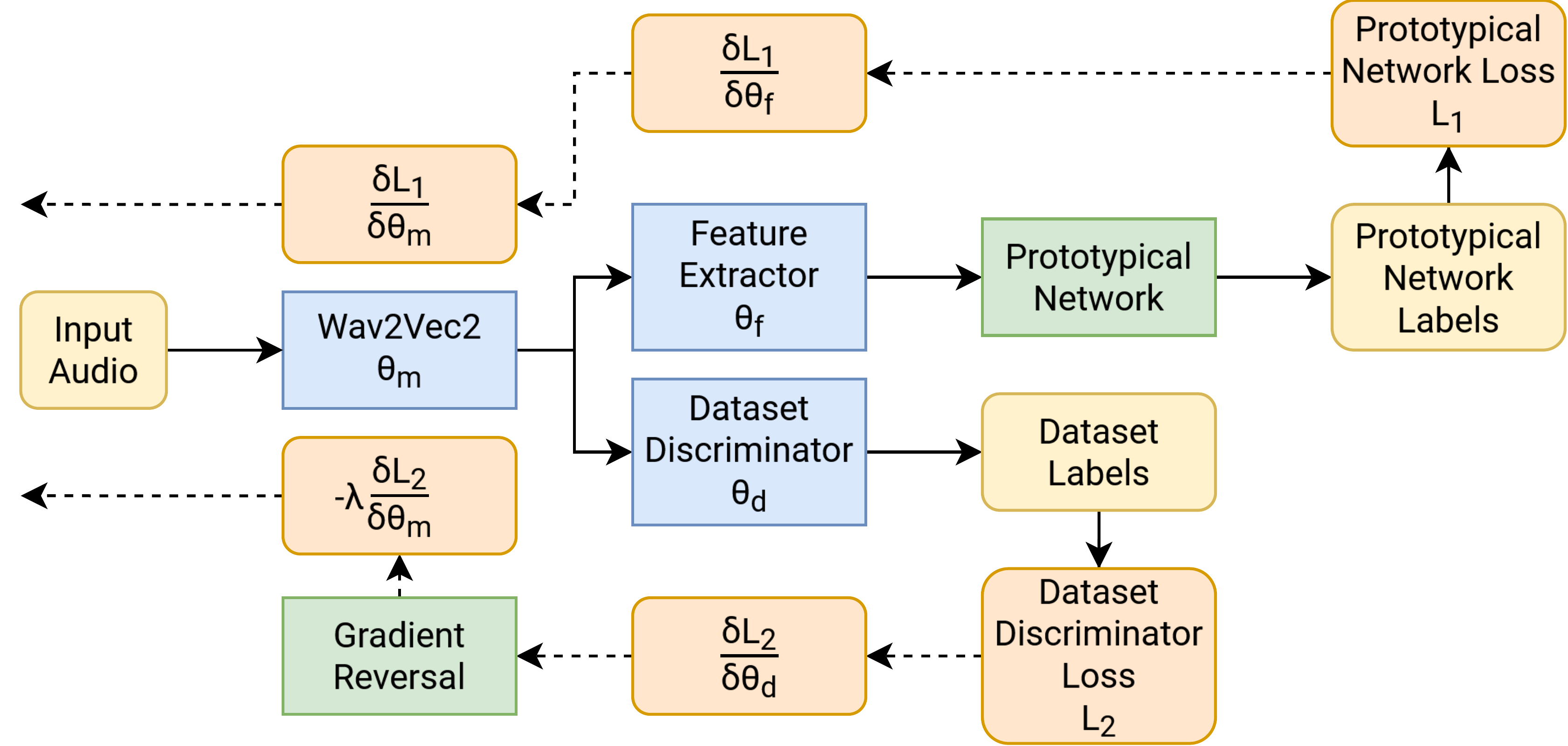}
\caption{The model architecture consists of the Wav2Vec2 XLS-R 300M backbone, the prototypical network on the top branch, and the domain adversarial dataset discriminator on the bottom branch. $\theta_m$, $\theta_f$, and $\theta_d$ represent the parameters for the embedding model, feature extractor, and dataset discriminator, respectively, and the dotted lines represent the gradient flows. $\lambda$ controls the domain adaptation influence.}
\label{fig:architecture}
\end{figure}

The overall architecture is shown in Figure \ref{fig:architecture}, employing several components. The backbone consists of the Wav2Vec2 XLS-R 300M model \cite{babu2021xlsr,baevski2020wav2vec}, which maps an audio input to a latent representation. Next, the feature extractor generates the embeddings used by the prototypical network. To modify the pre-trained backbone as little as possible, we do not use a parameter-less feature extractor to create the prototypical network embeddings. As Kumar et al. \cite{kumar2022fine} showed in their work, it is beneficial to only train the randomly initialized portion of a model, i.e., the feature extractor, before unfreezing the backbone and training the model as a whole. This approach minimizes propagating random noise into the backbone. Optionally, we attach a dataset discriminator to our model to implement domain adversarial training \cite{ganin2015unsupervised}.

\subsection{Feature Extractors}

We compare three feature extractor architectures to determine which approach maximizes the performance. Since the Wav2Vec2 XLS-R 300M model is not trained to receive a CLS token as input, we apply techniques that collapse the resulting variable-length embedding matrix depending on the input length into a fixed-size vector.

\textbf{Mean-FC} is one of the most straightforward available feature extractors. It averages the per-channel features into a 1024-dimensional vector, thus eliminating the dynamic component. It is followed by a fully connected layer, which further reduces the size of the embeddings and can be trained using linear probing.

\textbf{Lateral Inhibition} \cite{mitrofan-pais-2022-improving,pais-2022-racai} involves selectively disabling certain values, similar to attention. This layer follows Eq. \ref{eq:li}, where $\Diag(\cdot)$ takes as input a vector and outputs a diagonal matrix with the main diagonal equal to the input vector, $\ZeroDiag(\cdot)$ zeros the elements on the main diagonal of a matrix, $\Heaviside$ represents the Heaviside step function (i.e., it returns 0 for negative inputs, and 1 for positive inputs), and $W$ and $b$ represent trainable parameters. This lateral inhibition layer is followed by a fully connected layer, which reduces the embedding vector's dimensionality. Finally, a mean over the sequence length obtains a fixed-length embedding vector.

\begin{equation}
LI(x) = x \cdot \Diag(\Heaviside(x \cdot \ZeroDiag(W^T)+b))
\label{eq:li}
\end{equation}

\textbf{Gated Linear Unit} (GLU) \cite{dauphin2017language,shazeer2020glu} has been successfully used to reduce a variable-length matrix into a fixed-length vector \cite{raff2020classifying}. This layer is also reminiscent of attention; however, unlike lateral inhibition, it does not require the replacement of any gradient function. The implementation of this layer follows Eq. \ref{eq:glu}. The original implementation used fully connected layers instead of the convolutional layers $\Conv(\cdot)$. However, unlike the 1D convolutional layers we use in this case, those do not work for variable lengths. Therefore, the output of this layer is passed through a per-channel max function to reduce the variable sequence length, leaving us with a fixed-length vector.

\begin{equation}
GLU(x) = \Conv_A(x) \cdot \Sigmoid(\Conv_B(x))
\label{eq:glu}
\end{equation}

\subsection{Prototypical Network}

A prototypical network \cite{snell2017prototypical} is a parameter-less module that takes an initial set of embeddings in a latent feature space for support (i.e., labeled data) and a query (i.e., unlabeled data) set obtained through a feature extractor. It then computes a class prototype for each class in the support set. The class prototypes are calculated by averaging the embeddings for all $K$ examples of the same class. There are $N$-class prototypes in the context of $N$-way $K$-shot learning. Finally, examples in the query set are compared with each class prototype using a distance metric, like cosine similarity in our case, to determine the closest match, which determines the assigned label. The embedding function can be transfer-learned from another already-trained model. Fine-tuning consists of identifying the classes in the dataset, how close they are to each other using the distance metric, and maximizing the distance between different classes. In this way, we move the embeddings for different class examples far from one another and closer if they are from the same class.

\subsection{Training Procedure}

The training procedure follows the P\textbf{$>$}M\textbf{$>$}F stages \cite{hu2022pushing}: pre-training, meta-learning, and fine-tuning. We employ the pre-training checkpoint provided by the Hugging Face transformers library \cite{wolf2020transformers} for the pre-training stage.

\subsubsection{Meta-Learning.}

The general meta-learning procedure consists of supplying labeled examples, called the support set, and unlabeled examples, called the query set, on which we wish to perform inference. We use a prototypical network \cite{snell2017prototypical} as our meta-learning algorithm. We follow the indications in \cite{kumar2022fine} for the training by performing linear probing before training. As a result, we initially trained only our feature extractor for a few epochs before unfreezing the backbone (i.e., the Wav2Vec2 XLS-R 300M) and trained the entire model as a whole.

\subsubsection{Fine-Tuning.}

Similar to Hu et al. \cite{hu2022pushing}, we perform fine-tuning during meta-testing to adapt our model to unknown data distributions better. They showed that employing data augmentation in the context of the domain shift between training and testing plays a critical role in achieving good performance. In the speech emotion recognition setting, we use SpecAugment \cite{park2019specaugment} for the data augmentation strategy. Our approach is simpler than \cite{hu2022pushing}, which involved randomly selecting image transformations such as random brightness, contrast, translation, and horizontal flips. This is because SpecAugment masks time and frequency segments from the input audio spectrograms.

We experiment with using the support set both as support and query during fine-tuning, as suggested by \cite{hu2022pushing}. While this improves the performance, it does so at a high computational cost since we evaluate each support example twice because it has different SpecAugment masks each time. Our novel solution utilizes a subset of examples per class as support and the rest as a query. The disadvantage of this approach is it does not work for 1-shot learning. However, it runs twice as fast and achieves even better performance.

\subsubsection{Domain Adversarial Training.}

To further improve generalization, we incorporate the domain adversarial network (DANN) approach \cite{ganin2015unsupervised,ganin2016domain} through a dataset discriminator located after the Wav2Vec2 XLS-R embeddings (see Figure \ref{fig:architecture}). This discriminator predicts the dataset from which each sample came, as in a multi-class classification problem. Domain generalization is achieved by minimizing the classification loss for this problem for the parameters of the dataset discriminator while maximizing it for the parameters of the backbone. This adversarial objective is implemented by using the gradient reversal layer \cite{ganin2015unsupervised} before the discriminator, which reverses the gradient in the backpropagation step and scales it by a constant to avoid the divergence of the backbone's parameters.

\section{Experiments}

\subsection{Datasets}

\begin{table}[b]
    \centering
    \caption{Statistics of the datasets used for the three training tasks.}
    \begin{tabular}{*{6}{c|}c}
    \hline
         \multirow{2}{*}{Dataset Name} & \multicolumn{2}{c|}{Total} & \multicolumn{2}{c|}{Train} & \multicolumn{2}{c}{Test} \\ \cline{2-7}
          & \# Files & \# Tasks & \# Files & \# Tasks & \# Files & \# Tasks \\ \Xhline{4\arrayrulewidth}
         VoxCeleb2 & 1,092,009 & 5,994 & 982,750 & 5,395 & 109,259 & 599 \\ \hline
         Common Voice & 3,685,697 & 310 & 3,678,882 & 214 & 6,815 & 96 \\ \hline
         Reduced SER & 23,882 & 7 & 22,825 & 7 & 1,057 & 7 \\ \hline \Xhline{4\arrayrulewidth}
         All SER & 58,074 & 8 & 35,972 & 8 & 22,102 & 8 \\ \hline
    \end{tabular}
    \label{tab:datasets}
\end{table}

\begin{table}[!ht]
    \centering
    \caption{Breakdown of the Common Voice dataset adapted for accent recognition.}
    \begin{tabular}{c|c|c||c|c|c}
         \hline
         \multicolumn{6}{c}{Training Set} \\
         \hline
         Language & \# Accents & \# Files & Language & \# Accents & \# Files \\ \Xhline{4\arrayrulewidth}
         EN & 51 & 1,623,448 & IT & 7 & 168,114 \\ \hline
         FR & 26 & 663,563 & DE & 24 & 912,576\\ \hline
         ES & 12 & 469,985 & CA & 61 & 1,729,769 \\ \hline
         PT & 11 & 22,254 & RU & 6 & 27,803 \\ \hline
         RO & 1 & 5,206 & PL & 4 & 18,083 \\ \hline
         GL & 3 & 18,330 & SV-SE & 1 & 7,407\\ \hline
         NL & 4 & 60,940 & UK & 2 & 20,051\\ \hline
         BG & 1 & 3,385 &&&\\ \hline
         \multicolumn{6}{c}{} \\
         \hline
         \multicolumn{6}{c}{Test Set} \\ \hline
         Language & \# Accents & \# Files & Language & \# Accents & \# Files \\ \Xhline{4\arrayrulewidth}
         EN & 27 & 1,998 & IT & 8 & 586 \\ \hline
         FR & 7 & 412 & DE & 6 & 352 \\ \hline
         ES & 2 & 141 & CA & 33 & 2,322 \\ \hline
         PT & 10 & 749 & RU & 3 & 255  \\ \hline
    \end{tabular}
    \label{tab:cv_accents}
\end{table}

We train our model on person identification, accent recognition, and speech emotion recognition. Table \ref{tab:datasets} illustrates the differences between all the datasets described further.

\subsubsection{Person Identification} is done on the VoxCeleb2 dataset \cite{Chung18b}. The task is to correctly identify the person speaking in the test audio file, given 1 or 5 reference audio files. We chose this dataset because it features a larger speaker diversity of around 6,000 people from 145 nationalities.

\subsubsection{Accent Recognition} is done on the Common Voice dataset. To train on this task, we create the classification problem of determining the language-accent pair for an audio file. As shown in Table~\ref{tab:cv_accents}, we select 15 languages (i.e., English, Romanian, French, Italian, Spanish, Portuguese, Catalan, German, Swedish, Russian, Polish, Galician, Dutch, Ukrainian, and Bulgarian) from the Common Voice dataset, mainly spoken around Europe, and include the languages in the emotion recognition datasets. For this task, we freely sample an accent, such that it is possible that at one meta-training episode, all tasks contain different languages; thus, the problem degenerates into a language recognition problem. We have yet to explore the within-dataset sampling strategy to avoid this issue because most languages do not have an accent associated with them. We select only the language-accent pairs that contain at least 100 audio files and end up with over 200 tasks, the breakdown of which can be viewed in Table \ref{tab:cv_accents}.

\subsubsection{Speech Emotion Recognition} (SER) is done on a collection of datasets, illustrated in Table \ref{tab:emo_datasets}, where we use a total of 14 different datasets, among which 10 are used primarily for training, 2 for validation, and 2 for testing. The validation datasets are only used to tune the fine-tuning procedure performed during meta-testing. There is data in seven different languages, split as follows: English, Russian, Persian, Bengali, and German are used during training, German and Greek during validation, and Greek and Romanian during testing. The reasoning is to compare the performance in languages both included or not in the training set. Selecting which datasets to put in the validation and test sets is influenced by the amount of data each dataset contains since most data should be placed in the training set. The smallest datasets in our collection are EmoDB, EmoRO, AESI, and AESDD. EmoDB is a German dataset and putting it in the validation set allows us to test the performance of our approach on a different distribution while sharing a language present in the training set. AESI and AESDD are the only Greek datasets in our collection. We put one in the validation set and the other in the test set in a language not used during training. Lastly, EmoRO is the only dataset in Romanian we have, and placing it in the test set allows us to test on a language absent in the training and validation sets.

Finally, we also use a reduced subset of those datasets when benchmarking the dataset sampling techniques to speed up the training process. This is further referred to as the Reduced SER dataset (see Table \ref{tab:datasets}), which includes the RAVDESS \cite{livingstone2018ryerson}, DUSHA \cite{kondratenko2022large}, SUBESCO \cite{sultana2021sust}, EmoDB \cite{burkhardt2005database}, and EmoRO \cite{10314876} datasets. The complete collection of datasets is referred to as the All SER dataset.

\subsection{Data Preprocessing}

Due to computing limitations, we filter out all audio files in the SER datasets that are over 9.375s, or the equivalent of 150,000 waveform samples, knowing that Wav2Vec2 XLS-R 300M uses a sampling rate of 16kHz. This removes around 3\% of the total number of audio files (i.e., 1,800 files) or about 8.5\% of the total audio length (i.e., 3 hours). We do not truncate the data to avoid losing relevant information, such as intonation, at the end of the audio, which would lead to misclassification.

\subsection{Dataset Sampling}

We explore two dataset sampling techniques \cite{heggan2022metaaudio}: free dataset sampling and within-dataset sampling. The former requires sampling any task from any dataset, whereas the latter implies that only tasks from one dataset are present in a batch at any time. The implications are that with free dataset sampling, people speaking different languages will be sampled in the same batch, which can degenerate into a language classification problem instead of using emotional data for the prediction. In contrast, within-dataset sampling forces the network to use the emotional data instead of the language or other audio characteristics, such as the accent or background noise, since all audio in the same batch comes from the same dataset.

\begin{table}[!t]
    \centering
\caption{List of all datasets used for speech emotion recognition, along with the data split between the train, validation, and test sets.}
    \begin{tabular}{*{6}{c|}c}
    \hline
    \multirow{2}{*}{Dataset} & \multirow{2}{*}{Language} & \multirow{2}{*}{Hours of audio} & \multirow{2}{*}{\# Files} & \multicolumn{3}{c}{Used in} \\
    \cline{5-7}
    & & & & Train & Validation & Test \\
    \Xhline{4\arrayrulewidth}
    CREMA-D \cite{cao2014crema} & English & 5.26 & 7,442 & \checkmark & \checkmark & \checkmark \\
    \hline
    IEMOCAP \cite{busso2008iemocap} & English & 5.00 & 4,430 & \checkmark & \checkmark & \checkmark \\
    \hline
    MELD \cite{poria2019meld} & English & 11.22 & 13,427 & \checkmark & \checkmark & \checkmark \\
    \hline
    RAVDESS \cite{livingstone2018ryerson} & English & 1.28 & 1,248 & \checkmark & \checkmark & \checkmark \\
    \hline
    TESS \cite{E8H2MF_2020}& English & 1.60 & 2,800 & \checkmark & \checkmark & \checkmark \\
    \hline
    DUSHA \cite{kondratenko2022large} & Russian & 15.44 & 14,577 & \checkmark & \checkmark & \checkmark \\
    \hline
    RES-D \cite{Aniemore} & Russian & 1.48 & 1,037 & \checkmark & \checkmark & \checkmark \\
    \hline
    SHEMO \cite{MohamadNezami2019} & Persian & 2.91 & 2,864 & \checkmark & \checkmark & \checkmark \\
    \hline
    SUBESCO \cite{sultana2021sust} & Bengali & 7.83 & 7,000 & \checkmark & \checkmark & \checkmark \\
    \hline
    THORSTEN \cite{thorsten} & German & 1.64 & 1,500 & \checkmark & \checkmark & \checkmark \\
    \hline
    AESI \cite{chaspari2015development} & Greek & 0.46 & 696 & - & \checkmark & \checkmark \\
    \hline
    EmoDB \cite{burkhardt2005database} & German & 0.41 & 535 & - & \checkmark & \checkmark \\
    \hline
    AESDD \cite{vryzas2018speech} & Greek & 0.69 & 604 & - & - & \checkmark \\
    \hline
    EmoRO \cite{10314876} & Romanian & 0.18 & 522 & - & - & \checkmark \\
    \Xhline{4\arrayrulewidth}
    \multicolumn{2}{c|}{Training Set} & 33.78 & 35,972 & \checkmark & - & - \\
    \hline
    \multicolumn{2}{c|}{Validation Set} & 9.58 & 9,687 & - & \checkmark & - \\
    \hline
    \multicolumn{2}{c|}{Test Set} & 12.05 & 12,415 & - & - & \checkmark \\
    \Xhline{4\arrayrulewidth}
    \multicolumn{2}{c|}{All aggregated} & 55.40 & 58,074 & \checkmark & \checkmark & \checkmark \\
    \hline
\end{tabular}
\label{tab:emo_datasets}
\end{table}

\subsection{Hyperparameters}

We use a prototypical network embedding size of 256, which is the output size of the feature extractors. For the GLU feature extractor, we use two 1D convolutional layers with a stride of 1, a kernel size of 32, and a 1D dropout probability of 0.1. Unless otherwise specified, we use a dropout of 0.5 for all fully connected layers in the feature extractor. In the domain adversarial training setting, we use a GLU feature extractor with the same parameters but an output size equal to the number of training datasets. The domain adaptation parameter $\lambda$ is set to 0.01.

The training uses gradient checkpointing with a query size of 12 samples per class and 5 in the support set. We train using the Adam optimizer \cite{kingma2014adam} with a constant learning rate of $10^{-4}$ and gradient accumulation to perform a backpropagation step every 20 forward steps. We do linear probing for 5 epochs before unfreezing the backbone and training until overfitting, with a minimum validation loss often reached in 2-4 epochs. Each epoch consists of 1,000 randomly sampled batches, equivalent to 50 gradient descent steps.

\subsection{Evaluation}

We test 4-way classification in either 1-shot or 5-shot scenarios. It is noteworthy that random chance accuracy is 25\%. Table \ref{tab:emo_datasets} shows that the training, validation, and test sets contain data from various datasets. For each dataset that includes data in more than one set, we split the list of samples into disjoint sets so there is no data leakage between the splits. This is further helped by the fact that some datasets are only in some splits, such as AESI \cite{chaspari2015development} or EmoRO \cite{10314876} being used only in the test split.

Comparing meta-learning for speech emotion recognition against other approaches \cite{cai2020meta,chopra2021meta,feng2021few} is difficult since no standard training or testing methodology exists. Emotion datasets, not necessarily speech-based, generally contain several classes that may or may not match another dataset's emotions. There is a way to convert the multi-class emotion classification problem to binary classification by mapping each emotion by arousal and valence \cite{feraru2015cross}. However, this eliminates the granularity of the multi-class classification task.

\begin{table}[!htbp]
    \centering
    \caption{Results on the Reduced SER test sets when training on the accent recognition task, along with the results for training on the person identification task.}
    \begin{tabular}{c|c|c||c|c}
    \hline
        & \multicolumn{2}{c||}{4-way 1-shot} & \multicolumn{2}{c}{4-way 5-shot} \\ \hline
        Language & Person & Accent & Person & Accent \\
        & Identification & Recognition & Identification & Recognition \\ \Xhline{4\arrayrulewidth}
        EN & 34.25 & \textbf{38.92} & 43.40 & \textbf{48.10} \\ \hline
        RU & 32.05 & \textbf{31.58} & 35.30 & \textbf{40.00} \\ \hline
        DE & 43.42 & \textbf{51.18} & 57.53 & \textbf{64.60} \\ \hline
        RO & 29.67 & \textbf{30.10} & 33.50 & \textbf{36.90} \\ \Xhline{4\arrayrulewidth}
        Average & 34.85 & \textbf{37.95} & 42.43 & \textbf{47.40} \\ \hline
    \end{tabular}
    \label{tab:cv_vox_results}
\end{table}

\begin{table}[!htbp]
    \centering
    \caption{Results on the test sets using the free dataset sampling and within-dataset sampling strategies and number of examples per class provided. Results are for training on the reduced SER dataset.}
    \begin{tabular}{*{4}{c|}c}
    \hline
        & \multicolumn{2}{c|}{4-way 1-shot} & \multicolumn{2}{c}{4-way 5-shot} \\ \hline
        Language & Free Dataset & Within-Dataset & Free Dataset & Within-Dataset \\
        & Sampling& Sampling& Sampling& Sampling\\ \Xhline{4\arrayrulewidth}
        DE & 62.29 & \textbf{77.71} & 76.60 & \textbf{80.25} \\ \hline
        RO & 35.81 & \textbf{37.49} & 43.55 & \textbf{46.35} \\ \Xhline{4\arrayrulewidth}
        Average & 49.05 & \textbf{57.60} & 60.08 & \textbf{63.30} \\ \hline
    \end{tabular}
    \label{tab:emo_results}
\end{table}

\section{Results}

\subsection{Person Identification and Accent Recognition}

The results in Table \ref{tab:cv_vox_results} indicate that the accent recognition task on Common Voice is more effective than on VoxCeleb2. Despite that, even the 5-shot setting results only approach the performance obtained using the emotion datasets on the 1-shot setting, as seen in Table \ref{tab:emo_results}. This indicates the challenge of training a model on a generic or related task and having it generalize well on other tasks.

\subsection{Dataset Sampling}

Table \ref{tab:emo_results} shows that the dataset sampling technique affects performance. We notice that within-dataset sampling offers better performance than free dataset sampling. This may be because the sampling technique in testing is always within the dataset, and training in the same setting prepares the model better for this evaluation phase. This result contradicts the findings in \cite{heggan2022metaaudio}, indicating that the sampling technique depends on the task. The performance proves that training on SER data outperforms person identification and accent recognition; therefore, we use this training objective for the remainder of this work.

\begin{table}[!ht]
    \centering
    \caption{Accuracy (\%) for training on the full SER collection of datasets. }
    \begin{tabular}{*{3}{c|}c}
    \hline
    Dataset & Language & Validation Accuracy & Test Accuracy \\
    \Xhline{4\arrayrulewidth}
    CREMA-D \cite{cao2014crema} & English & 79.03 & 80.31 \\
    \hline
    IEMOCAP \cite{busso2008iemocap} & English & 61.60  & 61.53  \\
    \hline
    MELD \cite{poria2019meld} & English & 36.24  & 34.87  \\
    \hline
    RAVDESS \cite{livingstone2018ryerson} & English & 92.76  & 90.30 \\
    \hline
    TESS \cite{E8H2MF_2020}& English & 99.97  & 99.93  \\
    \hline
    DUSHA \cite{kondratenko2022large} & Russian & 65.95  & 66.54  \\
    \hline
    RES-D \cite{Aniemore} & Russian & 69.25  & 68.58  \\
    \hline
    SHEMO \cite{MohamadNezami2019} & Persian & 89.27  & 86.21  \\
    \hline
    SUBESCO \cite{sultana2021sust} & Bengali & 93.61  & 92.71  \\
    \hline
    THORSTEN \cite{thorsten} & German & 98.43  & 99.38  \\
    \hline
    AESI \cite{chaspari2015development} & Greek & 49.24  & 49.41  \\
    \hline
    EmoDB \cite{burkhardt2005database} & German & 91.24  & 92.86  \\
    \hline
    AESDD \cite{vryzas2018speech} & Greek & - & 73.41  \\
    \hline
    EmoRO \cite{10314876} & Romanian & - & 52.53  \\
    \Xhline{4\arrayrulewidth}
    \multicolumn{2}{c|}{Overall} & \textbf{76.35} & \textbf{74.34}  \\
    \hline
\end{tabular}
\label{tab:default_results}
\end{table}

\begin{table}[!t]
    \centering
    \caption{Accuracy (\%) of the three feature extractors along with DANN on the validation datasets. The mean accuracy refers to the average across all validation datasets.}
    \begin{tabular}{*{3}{c|}c}
    \hline
    Feature Extractor & Mean Accuracy & EmoDB Accuracy & AESI Accuracy \\ \Xhline{4\arrayrulewidth}
    Mean-FC & 77.74 & 91.38 & 50.93 \\ \hline
    Lateral Inhibition & 77.64 & 92.24 & 50.60 \\ \hline
    GLU & \textbf{78.15} & \textbf{93.45} & \textbf{56.70} \\ \hline
    GLU + DANN & 78.13 & 93.17 & 55.89 \\ \hline
\end{tabular}
\label{tab:fx}
\end{table}

\subsection{Feature Extractors}

The overall performance remains relatively the same when testing various feature extractors. However, there is a significant increase in performance for challenging datasets, such as AESI. Detailed results for a training run when using the GLU feature extractor are shown in Table \ref{tab:default_results}. The results of the ablation study for the feature extractor used are shown in Table \ref{tab:fx}. We observe that employing DANN decreases the overall performance slightly. This may be because it requires more careful hyperparameter tuning or harms the information stored in the pre-trained backbone. This finding aligns with Hu et al. \cite{hu2022pushing}. In either case, more research is required in this direction.

We notice high differences between the performances on datasets sharing the same language, ranging from 36.24\% on the MELD dataset to 92.76\% on the RAVDESS dataset, both in English. This can be explained by the intrinsic difficulty of each task, which is influenced by audio quality, the voiced text, speakers' accents or the ability to convey certain emotions.

\subsection{Fine-Tuning During Meta-Testing}

We initially perform a hyperparameter search over the desired number of optimizer steps and learning rates. For this round, we keep the inner support and query sets equal to the initial support set, with the only difference being the augmentations, as stated in the original work \cite{hu2022pushing}. The results are in Table \ref{tab:ft_initial}.

\begin{table}[!t]
    \caption{Accuracy (\%) for the hyperparameter search on the two validation datasets performed only on the validation splits. The inner support and query sets are augmentations of the support set. Bold indicates the highest overall score.}
    \label{tab:ft_initial}
    \begin{minipage}{.5\linewidth}
        \centering
        \begin{tabular}{*{4}{c|}c}
            \hline
            \multicolumn{5}{c}{EmoDB dataset} \\
            \hline
            \# & \multicolumn{4}{c}{Learning Rate} \\
            \cline{2-5}
            &&&&\\[-1em]
            Steps & $10^{-3}$ & $10^{-4}$ & $10^{-5}$ & $10^{-6}$ \\
            \Xhline{4\arrayrulewidth}
            0 & \multicolumn{4}{c}{91.24} \\
            \hline
            1 & \hspace{0.3em} 31.60 \hspace{0.3em} & \hspace{0.3em} 67.10 \hspace{0.3em} & \hspace{0.3em} 92.98 \hspace{0.3em} & \hspace{0.3em} 91.57 \hspace{0.3em} \\
            \hline
            3 & 26.03 & 78.30 & 92.52 & 91.90 \\
            \hline
            5 & 25.80 & 79.45 & 92.37 & 92.20 \\
            \hline
            10 & 25.88 & 79.12 & 92.42 & 92.83 \\
            \hline
            15 & 26.12 & 79.27 & 92.57 & 93.20 \\
            \hline
            20 & 25.62 & 78.37 & 92.43 & 93.20 \\
            \hline
            25 & 26.35 & 78.85 & 92.67 & \textbf{93.23} \\
            \hline
        \end{tabular}
    \end{minipage}%
    \begin{minipage}{.5\linewidth}
        \centering
        \begin{tabular}{*{4}{c|}c}
            \hline
            \multicolumn{5}{c}{AESI dataset} \\
            \hline
            \# & \multicolumn{4}{c}{Learning Rate} \\
            \cline{2-5}
            &&&&\\[-1em]
            Steps & $10^{-3}$ & $10^{-4}$ & $10^{-5}$ & $10^{-6}$ \\
            \Xhline{4\arrayrulewidth}
            0 & \multicolumn{4}{c}{49.24} \\
            \hline
            1 & \hspace{0.3em} 29.45 \hspace{0.3em} & \hspace{0.3em} 41.68 \hspace{0.3em} & \hspace{0.3em} 52.10 \hspace{0.3em} & \hspace{0.3em} 47.80 \hspace{0.3em} \\
            \hline
            3 & 25.15 & 46.03 & 53.67 & 48.53 \\
            \hline
            5 & 25.08 & 46.10 & 53.40 & 49.53 \\
            \hline
            10 & 25.18 & 48.75 & 54.10 & 51.10 \\
            \hline
            15 & 25.18 & 48.78 & 55.37 & 51.83 \\
            \hline
            20 & 25.27 & 48.07 & \textbf{56.72} & 52.40 \\
            \hline
            25 & 25.12 & 48.82 & 56.57 & 52.80 \\
            \hline
        \end{tabular}
    \end{minipage} 
\end{table}

\begin{table}[t]
    \caption{Accuracy (\%) on the hyperparameter search for the two validation datasets performed only on the validation splits. The inner support and query sets are disjoint sets of the initial support set, of varying lengths. For all experiments, the learning rate used was $10^{-5}$. Bold indicates the highest overall score.}
    \label{tab:ft_novel}
    \begin{minipage}{.5\linewidth}
        \centering
        \begin{tabular}{*{4}{c|}c}
            \hline
            \multicolumn{5}{c}{EmoDB dataset} \\
            \hline
            \# & \multicolumn{4}{c}{Support Size} \\
            \cline{2-5}
            Steps & 1 & 2 & 3 & 4 \\
            \Xhline{4\arrayrulewidth}
            1 & \hspace{0.3em} 91.43 \hspace{0.3em} & \hspace{0.3em} 92.07 \hspace{0.3em} & \hspace{0.3em} 92.57 \hspace{0.3em} & \hspace{0.3em} 92.43 \hspace{0.3em} \\
            \hline
            3 & 91.87 & 92.40 & 93.03 & 92.37 \\
            \hline
            5 & 91.47 & 92.40 & 93.10 & 92.43 \\
            \hline
            10 & 91.83 & 92.57 & \textbf{93.37} & 91.70 \\
            \hline
            15 & 92.63 & 92.30 & 92.93 & 91.37 \\
            \hline
            20 & 92.87 & 92.77 & 92.60 & 91.73 \\
            \hline
            25 & 93.23 & 92.93 & 92.93 & 91.93 \\
            \hline
        \end{tabular}
    \end{minipage}%
    \begin{minipage}{.5\linewidth}
        \centering
        \begin{tabular}{*{4}{c|}c}
            \hline
            \multicolumn{5}{c}{AESI dataset} \\
            \hline
            \# & \multicolumn{4}{c}{Support Size} \\
            \cline{2-5}
            Steps & 1 & 2 & 3 & 4 \\
            \Xhline{4\arrayrulewidth}
            1 & \hspace{0.3em} 49.20 \hspace{0.3em} & \hspace{0.3em} 50.40 \hspace{0.3em} & \hspace{0.3em} 50.43 \hspace{0.3em} & \hspace{0.3em} 48.87 \hspace{0.3em} \\
            \hline
            3 & 51.13 & 52.73 & 52.23 & 50.30 \\
            \hline
            5 & 52.87 & 53.93 & 53.43 & 52.07 \\
            \hline
            10 & 53.27 & 55.10 & 55.13 & 52.30 \\
            \hline
            15 & 54.57 & 56.03 & 55.67 & 52.90 \\
            \hline
            20 & 55.17 & 57.30 & 56.27 & 53.17 \\
            \hline
            25 & 56.17 & \textbf{57.80} & 56.67 & 53.40 \\
            \hline
        \end{tabular}
    \end{minipage} 
\end{table}

\begin{table}[t]
    \centering
    \caption{Test sets accuracy (\%) using the optimal fine-tuning hyperparameters.}
    \begin{tabular}{*{5}{c|}c}
    \hline
    \# Steps & EmoDB & AESI & AESDD & EmoRO & Avg. \\
    \Xhline{4\arrayrulewidth}
    0 & \hspace{0.3em} 92.86 \hspace{0.3em} & \hspace{0.3em} 49.41 \hspace{0.3em} & \hspace{0.3em} 73.41 \hspace{0.3em} & \hspace{0.3em} 52.53 \hspace{0.3em} & \hspace{0.3em} 67.05 \hspace{0.3em} \\
    \hline
    1 & 93.22 & 51.16 & 77.16 & 53.86 & 68.85 \\
    \hline
    3 & 93.30 & 53.48 & 80.72 & 54.16 & 70.42 \\
    \hline
    5 & 93.30 & 55.02 & 82.00 & 55.12 & 71.36 \\
    \hline
    10 & \textbf{93.92} & 56.52 & 83.24 & 55.70 & 72.34 \\
    \hline
    15 & 93.58 & 56.90 & \textbf{83.78} & 55.88 & 72.54 \\
    \hline
    20 & 93.58 & 57.38 & 83.70 & 56.06 & 72.68 \\
    \hline
    25 & \textbf{93.92} & \textbf{58.30} & 83.32 & \textbf{56.30} & \textbf{72.96} \\
    \hline
\end{tabular}
\label{tab:ft_eval_fin}
\end{table}

We reach the same conclusion as Hu et al. \cite{hu2022pushing}, where no unique set of hyperparameters yields the best results. For the optimal learning rates, we observe that the performance increases with the number of steps applied, which indicates that our model is undertrained. Due to the high computational cost of training a 300M parameter transformer, we set a maximum number of steps of 25. Regarding the learning rate, since we know the EmoRO dataset is a more difficult one with performances similar to AESI, we choose to continue with the best learning rate for AESI, which is $10^{-5}$.

For our next round of hyperparameter tuning, we deviate from the original algorithm and split the support set into two disjoint sets: the inner support set and the remaining samples representing the query set. At each step, new inner support and query sets are sampled. The results can be seen in Table \ref{tab:ft_novel}. A support size of 2 yields the best results on the AESI dataset, which we keep as the optimal configuration. We then evaluate the model's performance on the test set using these settings, yielding a performance shown in Table \ref{tab:ft_eval_fin}.

This process improves the performance on the EmoRO dataset by 4\%. We observe the same ascending trend on the AESI dataset, where the more fine-tuning steps we do, the better the performance. Again, the issue with computational resources arises since we need to perform gradient descent during meta-testing. Despite that, the only factor left is determining what trade-off between the number of steps and inference time is acceptable.


\section{Conclusions}

We explored various approaches to increasing performance in a low-data regime for speech emotion recognition, using either this task in the multilingual setting or surrogate tasks, namely person identification and accent recognition. We showed that the former approach performs best, especially when using in-dataset sampling. We have seen that multiple incremental improvements can make a significant difference when applied together. The choice of architecture for the feature extractor, coupled with fine-tuning during meta-testing brought our performance on the EmoRO dataset.

In future work, we propose to train a model to translate between emotional data. That is, to train a model to generate audio data with different emotions from a base emotion, possibly using CycleGAN \cite{zhu2017unpaired}. This would allow us to use the Common Voice dataset, which features data in many other languages and changes the emotion from neutral to any emotion we want. This way, we can train a model on more data in any possible language, not just the datasets already available.

\section*{Acknowledgements}
This work was supported by the National University of Science and Technology POLITEHNICA Bucharest through the PubArt program.

\bibliographystyle{splncs04}
\bibliography{mybibliography}

\end{document}